# Wasserstein GAN-Based Precipitation Downscaling with Optimal Transport for Enhancing Perceptual Realism

**Kenta Shiraishi[1], Yuka Muto[2], Atsushi Okazaki[2,3] and Shunji Kotsuki[2,3,4]**

[1]Graduate School of Science and Engineering, Chiba University, Chiba, Japan

[2]Center for Environmental Remote Sensing, Chiba University, Chiba, Japan

[3]Institute for Advanced Academic Research, Chiba University, Chiba, Japan

[4]Research Institute of Disaster Medicine, Chiba University, Chiba, Japan

Corresponding author: Shunji Kotsuki (shunji.kotsuki@chiba-u.jp)

**Key Points:**

- A Wasserstein Generative Adversarial Network (WGAN) was applied to precipitation downscaling with an optimal transport cost.

- The WGAN generated visually realistic fine-scale precipitation, despite slightly lower scores on conventional evaluation metrics.

- The trained critic of WGAN correlated well with human perceptual realism, and identified anomalies in both generated and reference data.

**Abstract**

High-resolution (HR) precipitation prediction is essential for reducing damage from stationary and localized heavy rainfall; however, HR precipitation forecasts using process-driven numerical weather prediction models remains challenging. This study proposes using Wasserstein Generative Adversarial Network (WGAN) to perform precipitation downscaling with an optimal transport cost. In contrast to a conventional neural network trained with mean squared error, the WGAN generated visually realistic precipitation fields with fine-scale structures even though the WGAN exhibited slightly lower performance on conventional evaluation metrics. The learned critic of WGAN correlated well with human perceptual realism. Case-based analysis revealed that large discrepancies in critic scores can help identify both unrealistic WGAN outputs and potential artifacts in the reference data. These findings suggest that the WGAN framework not only improves perceptual realism in precipitation downscaling but also offers a new perspective for evaluating and quality-controlling precipitation datasets.

**Plain Language Summary**

Providing high-resolution precipitation forecast is essential to reduce damage from stationary heavy rainfall. However, achieving the high-resolution forecast using traditional weather simulation models faces limitations, partly due to the computational power. This study used an artificial intelligence (AI) method called a Wasserstein Generative Adversarial Network to obtain high-resolution precipitation maps from low-resolution ones. This method captured small, detailed features of precipitation that are often lost in simpler AI models. Although the WGAN did not perform as well in conventional evaluation scores, the WGAN produced images that looked more realistic to the human eye. In addition, our results suggested that the WGAN is also applicable to detect potential quality issues in the data.

**1 Introduction**

The incidence of disasters caused by stationary heavy rainfall is increasing worldwide. Accurate and high-resolution (HR) precipitation forecasting is essential for reducing damage from stationary and localized heavy rainfall (e.g., Oizumi et al., 2020; Weisman et al., 2023). Conventionally, such HR precipitation forecasts have been achieved by increasing the spatial resolution of numerical weather prediction (NWP) models. However, this approach is computationally expensive and can become a bottleneck for frequent forecast updates. In fact, current NWP models still struggle to accurately simulate the initiation of band-shaped precipitation systems, and a positional error of approximately 25 km must be tolerated for forecasts to remain useful (Kato et al., 2018). In this study, we focused on artificial intelligence (AI) as an alternative approach to downscale coarse-resolution precipitation forecasts.

In the field of computer vision, various super-resolution (SR) techniques have been developed to downscale low-resolution (LR) images. These include convolutional neural network (CNN)-based methods, such as the super-resolution CNN (SRCNN; Dong et al., 2015) and Generative Adversarial Network for Super-Resolution (SRGAN; Ledig et al., 2017), as well as transformer-based methods such as SwinIR (Liang et al., 2021). Based on these successes, several studies have applied AI-based SR methods for precipitation downscaling. However, conventional

AI techniques can produce overly smooth and unrealistic precipitation and weather fields (e.g., Ravuri et al., 2021; Lam et al., 2023) mainly due to the choice of loss function. This issue is known as double penalty problem in precipitation evaluations (Michaelides, 2008). When using simple losses, such as mean squared error (MSE), weak and widespread precipitation forecast tend to be favored over more realistic but spatially shifted intense precipitation, because the former can yield lower MSE.

To address this issue, we propose a SR method for precipitation fields based on a deep learning model that uses the optimal transport distance as its loss function. The optimal transport distance has recently attracted a great deal of attention in climate and weather science (e.g., Vissio et al., 2020; Wan et al., 2023; Bocquet et al. 2024; Duc and Sawada 2024; Nishizawa 2024). However, standard formulations of optimal transport assume no bias between input and output distributions, and the mathematically exact solution is typically non-differentiable, making it difficult to use directly in neural network training. To overcome this, we used the Wasserstein Generative Adversarial Network (WGAN; Arjovsky et al., 2017), in which a generator that produces HR precipitation fields is trained together with a critic that learns the Wasserstein distance, a specific form of optimal transport distance. Here, the critic provides a differentiable approximation of the optimal transport cost, which the generator minimizes during training. The WGAN is a type of GAN that enables stable training using the Wasserstein distance and has shown promising results in image processing applications (Arjovsky et al., 2017; Gulrajani et al., 2017). We apply the WGAN to the SR of precipitation fields and compared its performance with conventional SRCNN. In addition to quantitatively evaluating the WGAN-based super-resolution, this study also investigates the characteristics of the trained critic and explores its potential applications in perceptual evaluation of downscaled precipitation data.

## 2 Methods and Data

### 2.1 Wasserstein Generative Adversarial Network (WGAN)

Let us first introduce the concept of the WGAN, whose schematic image is shown in Figure 1. The generator $G_\theta$ maps a LR precipitation input $z \sim p(z)$ to a HR output $\tilde{x} = G_\theta(z)$, where $z$ and $x$ are the LR and HR precipitation data, respectively. $\theta$ is the parameter of the generator. The generated HR data follow $\tilde{x} \sim \mathbb{P}_\theta$. The critic $F_\phi$ parameterized by $\phi$ receives both the real HR data $x \sim \mathbb{P}_r(x)$ and generated data $\tilde{x}$, and estimates the Wasserstein distance $W$ between the two distributions $\mathbb{P}_\theta$ and $\mathbb{P}_r$. The training step trains the generator and critic simultaneously so that the generator $G_\theta$ tries to reduce $W$, whereas the critic $F_\phi$ tries to maximize $W$ under the 1-Lipschitz constraint. Here, the 1-Lipschitz constraint ensures that the critic function does not change too rapidly with respect to its input. Consequently, the WGAN framework enables the generation of HR data from LR data with a reduction of $W$.

The WGAN is a framework for training generative models by minimizing the 1-Wasserstein distance (a.k.a. the Earth mover's distance), which is a form of optimal transport cost (Arjovsky et al., 2017). Mathematically, the distance between the real data distribution $\mathbb{P}_r$ and the generated data distribution $\mathbb{P}_\theta$ can be expressed via the Kantorovich–Rubinstein duality (Villani, 2009) as:

$$W(\mathbb{P}_r, \mathbb{P}_\theta) = \sup_{\|F_\phi\|_L \leq 1} \mathbb{E}_{x \sim \mathbb{P}_r}[F_\phi(x)] - \mathbb{E}_{\tilde{x} \sim \mathbb{P}_\theta}[F_\phi(\tilde{x})] \tag{1}$$

where the supremum is taken over the set of real-valued functions $F_\phi: x \to \mathbb{R}$ that satisfy the 1-Lipschitz condition. Here, $\|\cdot\|_L$ indicates the Lipschitz norm whose practical implementation is described in Section 2.3. Using this formulation, the training objective of the WGAN can be posed as the following mini-max optimization problem:

$$\min_\theta \max_{\phi: \|F_\phi\|_L \leq 1} \left( \mathbb{E}_{x \sim \mathbb{P}_r}[F_\phi(x)] - \mathbb{E}_{z \sim p(z)}[F_\phi(G_\theta(z))] \right). \tag{2}$$

This equation means that generator $G_\theta$ tries to minimize the critic-derived Wasserstein distance $W(\mathbb{P}_r, \mathbb{P}_\theta)$ whereas the critic $F_\phi$ tries to maximize the Wasserstein distance under the 1-Lipschitz constraint. To enforce the 1-Lipschitz constraint in practice, the original WGAN employed weight clipping. However, to improve gradient stability, the gradient penalty method is often adopted (Gulrajani et al., 2017); therefore, the gradient penalty was also applied in this study, as described in Section 2.3.

### 2.2 Data

We used radar–gauge merged precipitation analysis data provided by the Japan Meteorological Agency (Makihara et al., 1996) to evaluate the performance of WGAN in the precipitation downscaling. The precipitation analysis data provide hourly precipitation estimates with a spatial resolution of approximately 1 km × 1 km, by combining surface precipitation radar observations with approximately 1,300 rain gauge stations across Japan. We focused on a region of 128 km × 128 km (i.e., 128 × 128 pixels) centered over Chiba Prefecture, Japan, covering the period from 2006 to 2020. Here, the original 1 km × 1 km precipitation analysis data were used as the ground truth for evaluating generated precipitation fields. The corresponding LR inputs for the SR task were obtained by down-sampling the original data to a resolution of 4 km × 4 km.

Data from 2006–2014, 2015–2017, and 2018–2020 were used for training, validation, and test, respectively. We included only samples in which > 20 % of the pixels within the target region showed precipitation ≥ 0.4 mm hr$^{-1}$. This criterion was adopted to exclude scenes dominated by non-precipitating areas, thereby preventing the model from overfitting to no-rain patterns. Consequently, the data sizes of the training, validation, and test data sets were 15623, 5227, and 5950, respectively.

To improve the training efficiency and inference stability of the WGAN, we applied intensity normalization as follows. All precipitation values were linearly scaled to the interval [0, 1]. The upper threshold for scaling was set to 20 mm hr$^{-1}$, which covered approximately 99.8% of all data points. This avoided the problem of compressing low-intensity variability when normalizing based on the absolute maximum. The normalization was defined as:

$$\hat{R} = \min\left(\frac{R}{20}, 1\right) \tag{3}$$

where $R$ is the original precipitation intensity and $\hat{R}$ is the normalized value.

### 2.3 Experiments and Evaluation

This study discusses the downscaling performance of the WGAN in comparison to a conventional CNN-based deep neural network. As a baseline for comparison, we trained the

SRCNN (Dong et al., 2016) for precipitation downscaling. The SRCNN consists of a simple three-layer architecture, which is identical to the generator architecture used in the WGAN in this study. See Supplemental Figure S1 for more details of the network structures of SRCNN and WGAN. The loss function of the SRCNN is given by:

$$\mathcal{L}_{SRCNN} = \frac{1}{BN}\sum_{b=1}^{B}\sum_{k=1}^{N}\left(G_\theta\left(z_k^{(b)}\right) - x_k^{(b)}\right)^2, \quad (4)$$

where $B$ is the batch size, and was set be 512 in this study. $N$ is the number of image pixels (i.e., 128 × 128 = 16,384). $b$ and $k$ are the indexes of data in batches and pixels. On the other hand, the loss functions of the generator and critic in WGAN ($\mathcal{L}_{WGAN-G}$ and $\mathcal{L}_{WGAN-C}$) are given by:

$$\mathcal{L}_{WGAN-G} = \underbrace{-\frac{1}{B}\sum_{b=1}^{B}F_\phi\left(G_\theta(z^{(b)})\right)}_{Wasserstein\ distance} + \underbrace{\frac{\alpha}{BN}\sum_{b=1}^{B}\sum_{k=1}^{N}\left(G_\theta\left(z_k^{(b)}\right) - x_k^{(b)}\right)^2}_{MS}, \quad (5)$$

$$\mathcal{L}_{WGAN-C} = \underbrace{\frac{1}{B}\sum_{b=1}^{B}F_\phi(x^{(b)}) - \frac{1}{B}\sum_{b=1}^{B}F_\phi\left(G_\theta(z^{(b)})\right)}_{Wasserstein\ distance} + \underbrace{\frac{\lambda}{B}\sum_{b=1}^{B}\left(\left\|\nabla_{\hat{x}^{(b)}}F_\phi(\hat{x}^{(b)})\right\|_2 - 1\right)^2}_{1-Lipschitz\ constraint}. \quad (6)$$

Equation (5) indicates that the generator was trained to reduce the Wasserstein distance (the first term) with reduction of MSE (the second term) weighted by the tunable parameter $\alpha$, which was set to 10.0 based on preliminary investigations. Here, the first term of the Wasserstein distance $\frac{1}{B}\sum_{b=1}^{B}F_\phi(x^{(b)})$ is not included in the cost function because it is insensitive to $\theta$. The third term of Eq. (6) is the gradient penalty for enforcing the 1-Lipschitz constraint (Gulrajani et al., 2017) weighted by $\lambda$, which was set to 10.0 following Gulrajani et al. (2017). For the gradient penalty,

$$\hat{x}^{(b)} = \epsilon x^{(b)} + (1-\epsilon)G_\theta(z^{(b)}) \quad (7)$$

was used where $\epsilon \sim u(0,1)$ is the random coefficient.

For comprehensive evaluation of the performance of the generated precipitation images, we employed three commonly used metrics: root mean squared error (RMSE), the critical success index (CSI), and spatial frequency spectrum analysis. The CSI (also known as the thread score) is a metric that simultaneously evaluates precision and recall in event detection, defined by:

$$CSI = \frac{TP}{TP+FP+FN}, \quad (8)$$

where abbreviations are true positives (TP), false positives (FP), and false negatives (FN), respectively. In this study, threshold values of 10 and 15 mm hr$^{-1}$ were used. CSI quantifies how accurately the model reproduces meteorologically significant precipitation events. In addition, we applied a two-dimensional fast Fourier transform (FFT) to both the reference and generated images to evaluate the ability of the model to reconstruct fine-scale structures.

## 3 Results and Discussion

### 3.1 Evaluation of Inferences

Figure 2 presents an example of downscaled precipitation at 16:00 UTC on 5 July 2020. Precipitation fields in the SRCNN output were excessively smoothed, which is a typical characteristics of machine learning models that were only trained using MSE for precipitation inference. The WGAN, on the other hand, replicated localized high-intensity pixels and sharper edges that were very similar to the original HR data. This implied that fine-scale precipitation patterns were better preserved by WGAN. Both SRCNN and WGAN outperformed the LR input, suggesting better detection of areas with heavy precipitation in this example. A visual comparison revealed that WGAN more accurately depicted the realistic structure of precipitation, even though SRCNN performed better in terms of RMSE and CSI. This highlighted the limitation of depending only on traditional metrics such as RMSE and CSI for precipitation evaluations.

Table 1 presents a summary of the RMSE and CSI computed over test data. As anticipated, the RMSE of 0.273 mm hr$^{-1}$ for SRCNN, which directly minimized the MSE (Eq. 4), was lower than that of 0.292 mm hr$^{-1}$ for WGAN. SRCNN also performed better than WGAN in terms of the CSI. It can be said that SRCNN consistently outperformed WGAN in both pixel-wise accuracy and event-based detection.

Although SRCNN performed better than WGAN in quantitative measures (i.e., RMSE and CSI), there was a noticeable difference between their spectral properties (Figure 3). The SRCNN output's power spectrum rapidly decayed beyond about 20 pixels$^{-1}$, indicating a significant loss of high-frequency components that led to excessively smoothed precipitation fields. The WGAN output, on the other hand, closely matched the HR data spectrum and maintained spectral energy over a wider frequency range including fine-scale structures (20–60 pixel$^{-1}$). This implied a basic difference in modeling behavior: WGAN prioritized the reconstruction of spatial structure, producing more visually realistic precipitation patterns despite performing worse on traditional metrics than SRCNN. The WGAN prioritizes the reconstruction of fine-scale structures in a way that aligns more closely with human perception through adversarial training with a trained critic. This leads to sharper, more detailed images that capture localized high-intensity features, even if their exact placement can differ slightly from the reference.

**3.2 Evaluation of the Trained Critic**

Here we investigate the characteristic of the trained critic of WGAN by computing critic scores. Figure 4 (a) compares critic scores for HR, LR, SRCNN and WGAN data over test data. Note that critic scores of HR and WGAN data correspond to $F_\phi(x)$ and $F_\phi(G_\theta(z))$, meaning that the generator tries to reduce their expected difference by Eq. (2). Comparing the HR and LR, we observe that the HR data generally received higher critic scores. The SRCNN output showed only slight difference w.r.t. the LR input, suggesting that the critic did not perceive a substantial improvement. In contrast, the WGAN output showed a distribution that shifted slightly closer to that of the HR data.

Figure 4 (b) displays the differences in critic scores with respect to the HR data. Both SRCNN and WGAN underestimated the critic score compared to the HR data; however, WGAN exhibited a reduced underestimation bias relative to SRCNN. In the WGAN framework, the generator tried to decrease this difference under a fixed critic (Eq. 5), while the critic is trained to increase the gap to obtain a function that better distinguishes real HR data from generated data (Eq. 6). That is, the more negative (positive) the critic difference, the more likely the generator output was to be considered a success (failure). Therefore, we examine three representative cases where the critic difference was significantly positive or negative.

Figure 5 shows three cases with the largest negative critic differences between the HR and WGAN outputs $(F_\phi(x) - F_\phi(G_\theta(z)))$ = -1.64, -1.61 and -1.50, respectively). Interestingly, all three cases corresponded to line-shaped precipitation events. While SRCNN only produced blurred precipitation patterns, WGAN successfully reconstructed sharp and well-defined precipitation structures. These results demonstrated that large negative critic differences often corresponded to visually superior reconstructions, suggesting that the trained critic aligned well with human perception. Although SRCNN exhibited better RMSE and CSI values in these cases, its outputs failed to capture fine-scale detail.

Alternatively, Figure 6 shows three cases with the largest positive critic differences between the HR and WGAN outputs $(F_\phi(x) - F_\phi(G_\theta(z)))$ = +2.03, +1.67 and +1.63, respectively), indicating that the generator failed to produce perceptually convincing outputs. The first two examples exhibited visibly unnatural precipitation patterns in the WGAN outputs as indicated by red-colored rectangles (Figures 6 d-1 and d-2), suggesting that the trained critic may reflect perceptual judgments aligned with human visual assessment. The output from the WGAN lacked the speckled patterns observed in the HR data and instead exhibited unnatural, overly connected structures. On the other hand, no similarly unnatural patterns were found in the third-case generated data, shown in Figure 6 (d-3). However, in this case, the HR data exhibited an unusual precipitation pattern that appears to be radar clutter. When examining a wider area, we found other instances of unnatural, rectangular-shaped precipitation patterns at the same timestamp (Supplemental Figure S2). This example suggests that large critic differences may help identify unrealistic precipitation patterns embedded within the HR data itself. In other words, the critic may have potential applications in data quality control.

**3.3 Application to Coarse-Resolution Input Data**

Here, we qualitatively evaluated the sensitivity of model performance to the spatial resolution of input data. Unlike Section 3.1, where we used inputs down-sampled to 32 × 32 pixels, we retrained both the SRCNN and WGAN models using coarser inputs at a resolution of 16 × 16 pixels. Figure 7 presents the downscaling results for the same case as shown in Figure 2 (16:00 UTC, 5 July 2020).

It became significantly more difficult to visually identify the two diagonal precipitation bands in the input LR data (Figures 7 b and f). The SRCNN output tended to become even more spatially smoothed, further blurring fine-scale features and failing to recover the distinct precipitation structures. In contrast, the WGAN output showed the two diagonal bands clearly and recovered localized high-intensity features. Remarkably, the WGAN even inferred precipitation intensities stronger than those in the input data, demonstrating its ability to reconstruct realistic fine-scale structures despite the limited spatial information available. This behavior was notably different from the SRCNN.

These results highlighted the potential of adversarially trained models such as WGAN to extrapolate plausible HR features from coarse inputs, an ability that may be particularly valuable when only LR precipitation observations or forecasts are available.

## 4 Summary

This study proposed a SR approach for precipitation downscaling based on the WGAN, which enhances the realism of reconstructed precipitation patterns by utilizing optimal transport cost. The WGAN model was created to maintain both fine-scale spatial structures and quantitative consistency by combining adversarial training with MSE loss.

We used RMSE, CSI, and spectral analysis to compare the performance of WGAN with a baseline SRCNN. The WGAN preserved high-frequency components that are frequently lost in MSE-optimized SRCNN, resulting in more visually realistic precipitation fields, even though the SRCNN performed better in RMSE and CSI. This highlighted the limitations of standard evaluation metrics in capturing perceptual quality and spatial structure. Furthermore, we discovered that WGAN was more robust in low-resolution settings than SRCNN when it came to reconstructing sharp precipitation patterns from coarse inputs. In addition, analysis of the WGAN critic revealed that critic scores aligned well with human perception: large negative critic differences indicate realistic reconstructions, while large positive differences highlight perceptual failures. Our experiments suggested the critic's potential for both quality assessment and data quality control.

Among generative methods, GANs have been used to generate realistic images (e.g., Ravuri et al., 2021). However, unstable training is a common problem with GANs, which we also faced in our experimental setup. In contrast, WGAN offers a more stable training process by replacing the Jensen–Shannon divergence with the Wasserstein distance, which provides smoother gradients even when the generated and real distributions do not overlap. This makes WGAN a practically advantageous method for SR tasks in precipitation downscaling.

Overall, this study demonstrated that incorporating optimal transport into the SR of precipitation fields can enhance the visual and structural quality of outputs. In future work, we plan to apply the WGAN as a post-processing tool for precipitation forecasts by NWP models. In addition, applying the trained critic to evaluate the characteristics of precipitation products derived from NWP outputs or satellite observations presents a promising new direction. The trained critic could also be used as a loss function for other machine learning tasks beyond SR, such as for precipitation prediction.

**Acknowledgments**

This study was partly supported by the JST Moonshot R&D (JPMJMS2389), the Japan Aerospace Exploration Agency (JAXA) Precipitation Measuring Mission (PMM, ER4GPF019), the Japan Society for the Promotion of Science (JSPS) KAKENHI grants JP21H04571, JP21H05002, JP22K18821,JP 25K17687, JP25H00752, and the IAAR Research Support Program and VL Program of Chiba University.

**Conflict of Interest**

The authors declare that they have no conflict of interest.

**Open Research**

The data archiving is underway, and will be opened on zenodo (https://zenodo.org/) by the time of publication. In addition, all of the data and codes used in this study are stored for 5 years at Chiba University.

**Table 1.** Statistical comparison of downscaled precipitation scores, root mean squared error (RMSE), and critical success index (CSI), for SRCNN and WGAN averaged over test data.

| Method | RMSE (mm hr$^{-1}$) | CSI (10 mm hr$^{-1}$) | CSI (15 mm hr$^{-1}$) |
|---|---|---|---|
| LR (input) | 0.362 | 0.437 | 0.387 |
| SRCNN (inference) | 0.273 | 0.467 | 0.416 |
| WGAN (inference) | 0.292 | 0.448 | 0.387 |

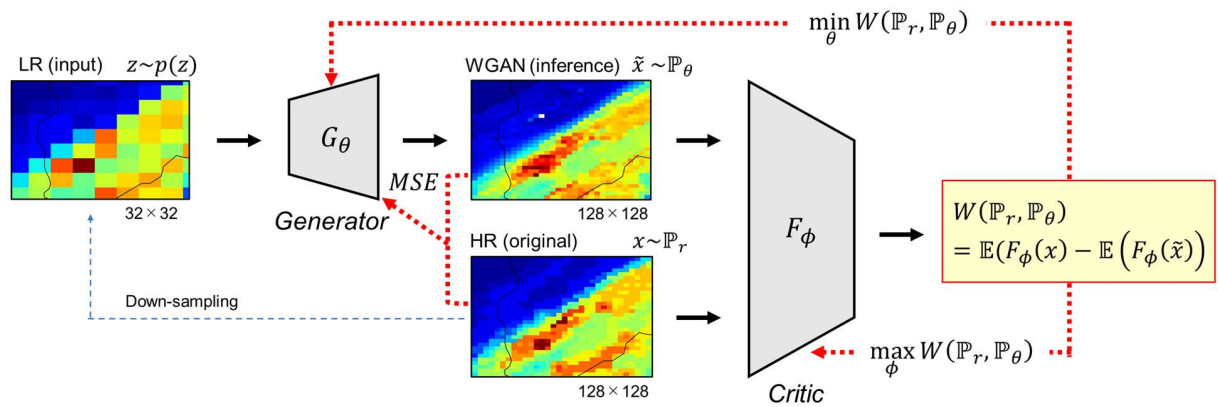

**Figure 1**. The schematic representation of the Wasserstein Generative Adversarial Network (WGAN) applied for precipitation downscaling. This schematic image describes the training steps to update the model parameters of the generator and critic from the original high-resolution images.

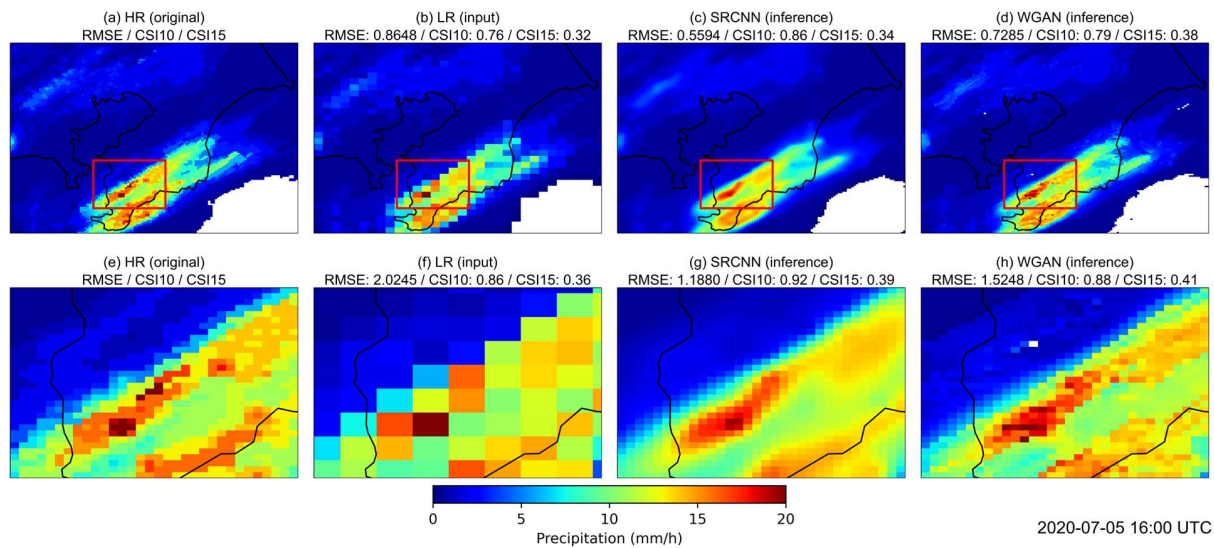

**Figure 2**. An example precipitation field (mm hr$^{-1}$) at 16:00 UTC on 5 July 2020, for (a, e) original high-resolution (HR) data (128 × 128), (b, f) input low-resolution (LR) data (32 × 32), (c, g) inference of SRCNN, and (d, h) inference of WGAN. (a–d) Full domain. (e–h) Close-up views of the red-colored rectangles indicated in (a–d). Root mean square error (mm hr$^{-1}$) and critical success index (CSI) with thresholds (10 and 15 mm hr$^{-1}$) were computed against the HR data.

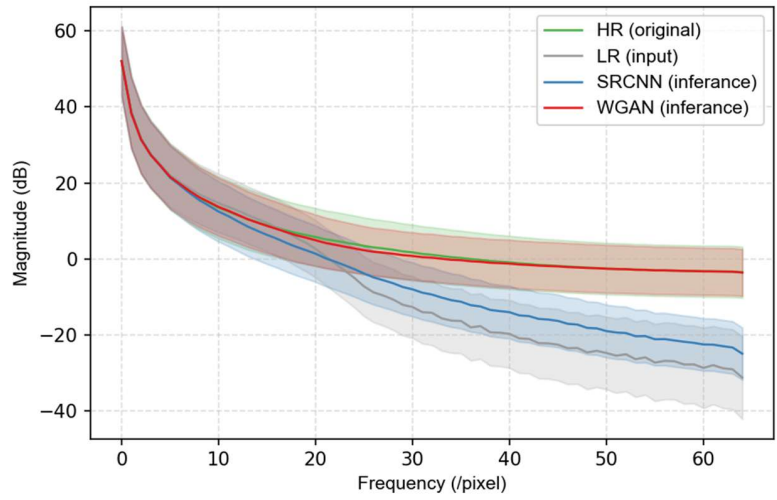

**Figure 3**. Power spectrum comparison of precipitation fields averaged over test data. The gray line represents the LR input, the green line shows the original HR data, the blue line shows the inference by SRCNN, and the red line shows the inference by WGAN. Lines and shading represent the mean ± σ of the power spectrum.

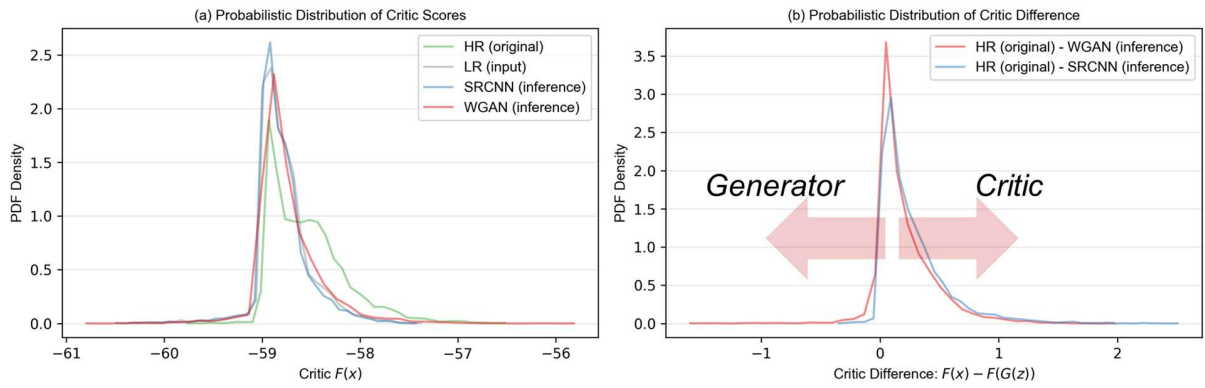

**Figure 4**. Probabilistic distributions of (a) critic scores and (b) critic difference over test data. (a) shows critic scores for HR (green line), LR (gray line), SRCNN (blue line) and WGAN (red line). (b) shows the critic difference between HR – WGAN (red line) and HR – SRCNN (blue line).

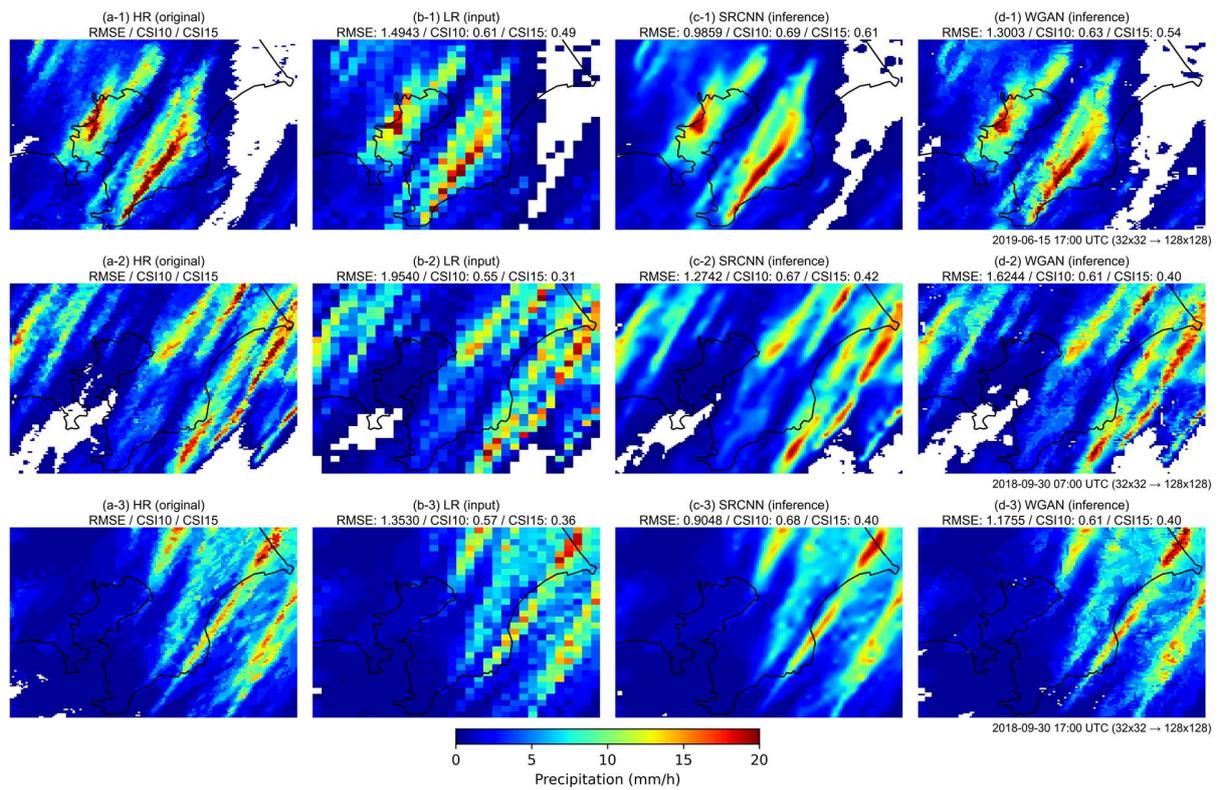

**Figure 5**. Similar to Figures 2 (a–d), but showing the three cases with the largest negative critic differences between the HR and WGAN outputs. Three cases correspond to 17:00 UTC on 15 June, 2019, 07:00 UTC on 30 September, 2018, and 17:00 UTC on 30 September, 2018.

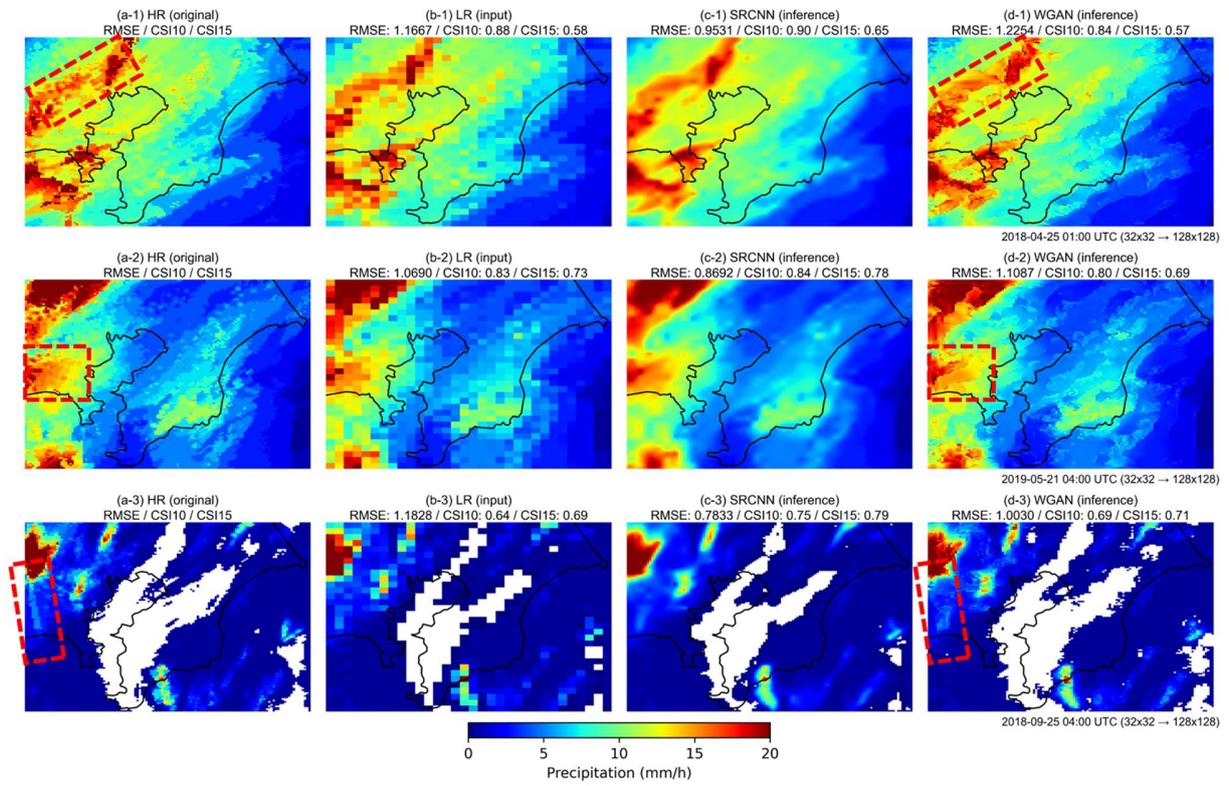

**Figure 6**. Similar to Figure 5, but showing the three cases with the largest positive critic differences between the HR and WGAN outputs. Three cases correspond to 01:00 UTC on 25 April, 2018, 04:00 UTC on 21 May, 2019, and 04:00 UTC on 25 September, 2018.

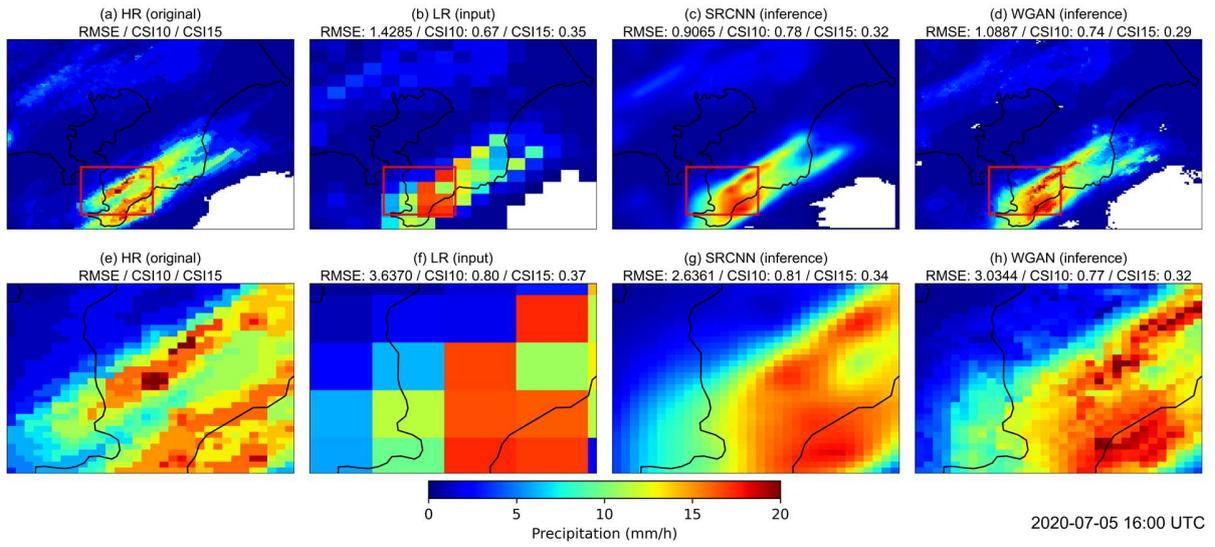

**Figure 7**. Similar to Figure 2, but for downscaling from coarser LR input (16 × 16 pixels).

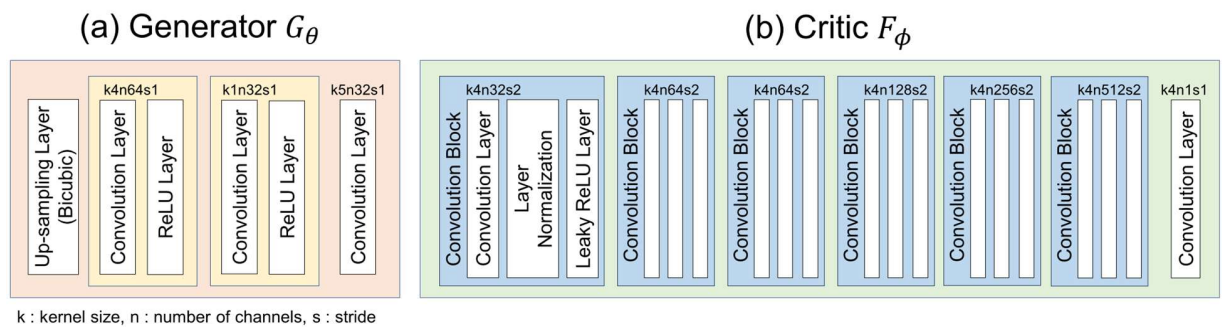

**Figure S1**. The network structure of the generator and critic of the WGAN used in this study. The SRCNN is identical to the architecture of WGAN's generator.

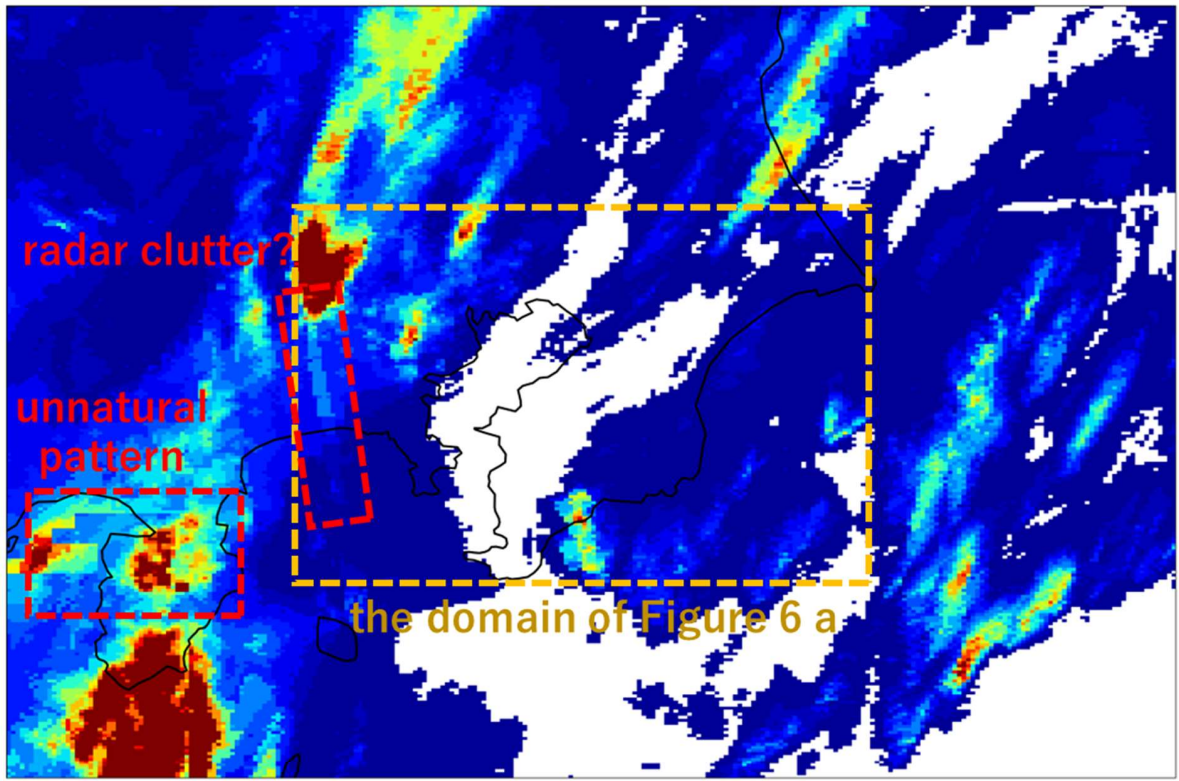

**Figure S2**. Same as Figure 6a but over a wider domain. Red rectangles highlight unnatural precipitation patterns in the data, while the orange rectangle indicates the area shown in Figure 6a.